# In-depth Question classification using Convolutional Neural Networks

**Prudhvi Raj Dachapally**[1,*] **and Srikanth Ramanam**[1,**]

[1]*School of Informatics and Computing, Bloomington, IN 47408, U.S.A.*
[*]*Corresponding authors: prudacha@indiana.edu*
[**]*Corresponding authors: srikrama@iu.edu*

**Answering systems have been one of the major areas of research in the field of Machine Learning and Natural Language Processing. Our goal in this work is to simplify the problem classifying the questions through a topic-oriented two-tier Convolutional Neural Network.**

## CONTENTS

## 1. INTRODUCTION

Convolutional neural networks for computer vision are fairly intuitive. In a typical CNN used in image classification, the first layers learn edges, and the following layers learn some filters that can identify an object. But CNNs for Natural Language Processing are not used often and are not completely intuitive. We have a good idea about what the convolution filters learn for the task of text classification, and to that, we propose a neural network structure that will be able to give good results in less time.

We will be using convolutional neural networks to predict the primary or broader topic of a question, and then use separate networks for each of these predicted topics to accurately classify their sub-topics.

## 2. RELATED WORK

The problem of question classification has been studied before but most of them are domain specific or restricted to high-level classification.

Anbuselvan et al. (2015) [1] proposed an SVM based method for the same task. The question is first parsed and tokenized, parts-of-speech are tagged, stop-words are removed, data is stemmed, and a lot of features are extracted. Feature selection steps are done before actually passing the data into a support vector machine for training. The same preprocessing is done for test questions as well, which might be time consuming to get results in real time.

Marco Pota et al. (2015) [2] propose a feature-based method where features related to specific subset of questions like wh-words, how – all/some words, head-verbs, and various other features like these were extracted from the texts before passing into a classifier. The results from this paper will be used as a baseline for comparing our sub-category results.

CNNs for NLP were used in a few works before. Collobert et al. (2011) [3] first proposed the idea of convolutional neural network structure which includes lookup tables, and hard hyperbolic tangents. Kalchbrenner et al. (2014) [4] proposed a simplified version of Collobert's network that was used to classify questions and Twitter sentiment. They used the concept of k-max pooling, which we adapted this method in our model. Yoon Kim (2014) [5] expanded on Kalchbrenner's work to add various machine learning strategies like regularization to make the network perform better.

## 3. CNN FOR NATURAL LANGUAGE PROCESSING

Since we are working with text, we need some methodology to convert these rawtexts into numerical representation.

### 3.1. Vector Representations of Text

The naïve approach is one-hot encoding, but as the size of the vocabulary in the dataset increases, this representation becomes bad. Another approach is to use a bag of words representation based on the corpus. But since we want our model to be not restricted to a specific domain or corpus, we used pre-trained word vectors like word2vec [6] and GloVe [7]. These models are trained on different datasets to create vector spaces with words positioned based on their semantics. These vector representations can be leveraged to learn the weights that work with related phrases accurately in classification tasks.

### 3.2. Convolutions on Text Data

Each sentence can consist of $m$ words. And each word is represented as a $d$-dimensional vector. So, a box (square) kernel would not be useful in this context. Therefore, we use something called a wide convolution. A wide convolution is represented as a $n \times d$ kernel, where $n$ is the number of words captured at a time, and $d$ is the dimensionality of the word vector. For example, if the height of the kernel is 2, then that filter captures all the bigrams sequentially in a sentence. Consider the case of general sentiment analysis. If the sentence is *"I am not very much interested"*, the model should classify this as a negative sentence. If we deconstruct this into bi-grams, phrases decoded will be like *'very much'*, and *'much interested'*. This is a positive phrase. But, if we consider tri-grams, the deconstructed phrases will be *'not very much'*, and *'very much interested'*. The latter phrase still possesses a positive connotation. Once we consider quad-grams, it deconstructs the sentence into phrases like *'am not very much'* and *'not very much interested'*. These phrases can help in determining that the sentence is a negative sentence.

For a CNN, given a good number of training examples, the network will be able to learn various filters that activate according to the phrases found in the input question. For this task, our intuition is that the bi-gram filters learn question openers like *where is, what are, who is, which is, who named* etc.. As the height of kernel increases, the network will be able to learn phrases of various lengths that can help it classify the intent of a given question.

## 4. PROPOSED SOLUTION

Our main idea in this project is to expand upon the existing work to create a two tier CNN that classifies question into their main and sub-categories. Since the argument is that convolutions are very fast, instead of creating a single network that can classify an example into 50 classes, we create separate network for each primary class, and this helps in giving the secondary layer CNN some prior information about the primary category.

The proposed architecture for the convolutional neural network consists of one convolutional layer that learns a number of filters for given heights. In this network, we capture from bi-grams to pent-grams. This helps us learn the intent of the question to a greater extent. For example, as said earlier, bi-grams capture the question starters, which is useful for predicting the broad categories in the first tier. Tri-grams to pent-grams are important for acquiring the actual intent of the question.

Next we added a k-max pooling layer (Kalchbrenner et al., 2014). We used 2-max pooling for our network to accumulate more information from the convolution filters. Then we merge all these max-pooled outputs to form a fully-connected layer. CNNs tend to work better when more fully-connected layers are added at the end before the output softmax layer [8], [9]. Therefore, we add two layers with N and N/2 hidden nodes with hyperbolic tangents as their activation functions. Dropout of 0.5 was used in these two layers to avoid overfitting while training. Fig 1 shows the structure of CNN we developed for question classification.

We are using two tiers of CNNs to classify the questions at different levels- main and sub-categories. An overview of this two-tier CNN can be found in the figure 2. The questions classified into their main categories by tier 1 CNN are directed to the appropriate CNN in tier 2 for determining their sub-category.

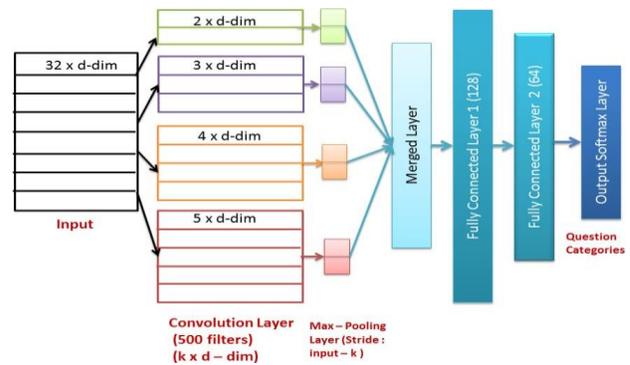

**Fig. 1.** Proposed CNN Architecture

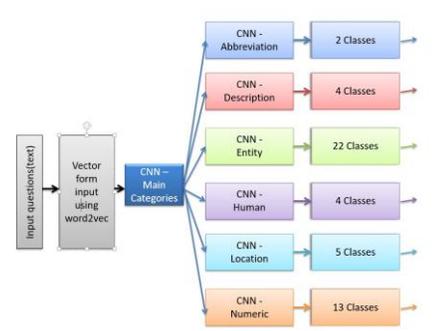

**Fig. 2.** Overview of the 2-tier CNN

## 5. DATA

The data used for training was question classification dataset by University of Illinois, Urbana Campaign. This has 5452 questions labeled as six primary categories and 50 sub-categories. The main categories and their corresponding number of sub-categories are listed in the table 1.

For testing, we used two different datasets; the TREC dataset with 500 questions by UIUC [10], and a manually collected dataset of 115 questions from Quora website [11]. The questions in Quora datset were randomly selected from various topics to ensure a good distribution across the several categories we are training our model on. These questions were manually labeled with appropriate main and sub categories.

| Main Category | Number of Sub-categories |
|---|---|
| **Abbreviation** | 2 |
| **Entity** | 22 |
| **Description** | 4 |
| **Human** | 4 |
| **Location** | 5 |
| **Numeric** | 13 |

**Table 1.** Number of sub categories in each category

| Coarse classes | Fine classes |
|---|---|
| ABBREVIATION | abbreviation, expression |
| ENTITY | animal, body, colour, creative, currency, disease, event, food, instrument, language, letter, other, plant, product, religion, sport, substance, symbol, technique, term, vehicle, word |
| DESCRIPTION | definition, description, manner, reason |
| HUMAN | group, individual, title, description |
| LOCATION | city, country, mountain, state, other |
| NUMERIC | code, count, date, distance, money, order, period, percent, speed, temperature, size, weight, other |

**Fig. 3.** Subcategories in each category. Image source: [2]

| Dataset | Sample Question |
|---|---|
| TREC | What do you call a newborn kangaroo? |
| Quora | What are some good Python tasks for a beginner of big data analysis? |

**Table 2.** Example questions

## 6. EXPERIMENTS

We performed three different experiments on the same training data set. The data used for training was question classification dataset[10] by University of Illinois, Urbana Campaign. In the first set of experiments, we used word2vec for vector representation of words. We used the Word2vec [6] model trained on Google News dataset. Each word is represented as a unique 300-dimensional vector using word2vec. For the second set, we used GloVe which was trained on Wikipedia and Gigaword 5. Both of these represent each word as a 300 dimensional vector.

To test the robustness of both the models, we tested this model on a new dataset compiled by us, with questions from Quora.

## 7. RESULTS

The results for the word2vec model are shown in the table below. Accuracy metric used here is the number of true positives divided by the total number of examples. The secondary level accuracy represents the sum of true positives divided by total number of inputs over all subcategories of all main categories.

For the second set, we used GloVe representations to train the model. These representations were giving an accuracy of 84.2% (421/500) for the primary categories on the TREC test data. But when we trained the subcategory CNNs with GloVe, the validation accuracy was comparable to word2vec. Therefore, instead of using GloVe channeled primary CNN, we used the previous word2vec channeled CNN as the first tier. The comparison of individual results between the sub-category accuracies, for both word2vec and GloVe models, are shown in the table 3. We observed that the accuracy was for categories with larger number of sub-categories. For example, low accuracies of 77.5% and 81.9% were observed for Entity with 22 sub-categories.

| Word2vec and GloVe sub-category results comparison | | | |
|---|---|---|---|
| Class | Entries | Word2vec accuracy | GloVe accuracy |
| Abbreviation | 9 | 100% | 100% |
| Description | 138 | 100% | 97.1% |
| Entity | 94 | 77.5% | 81.9% |
| Human | 65 | 98.46% | 98.46% |
| Location | 81 | 95% | 92.6% |
| Numeric | 141 | 92.9% | 93.61% |

**Table 3.** Results achieved through the two-tier CNN model

The comparison of our results with related previous works are shown in the 4.

| Comparison of Results with previous works | | |
|---|---|---|
| Model | Main category Accuracy | Sub category Accuracy |
| Pota et al. (2015) [2] | 89.6% | 82.0% |
| Yoon Kim (2014) [5] | 92.8% | - |
| Kalchbrenner et al. (2014) [4] | 93.0% | - |
| Our model (word2vec-word2vec) | 93.4% | **87.4%** |
| Our model(word2vec-Glove) | 93.4% | **87.2%** |
| Anbuselvan et al. (2015) [1] | **95%** | - |

**Table 4.** Comparing our results on TREC dataset with previous works

Next we tested our models with the dataset we compiled from Quora. The results for the Quora dataset are shown in the table 5. For the same dataset, the accuracy percentage of the sub categories, given that the main categories are correctly predicted, is 84.6% for word2vec–word2vec model.

| Model | Main category Accuracy | Sub category Accuracy |
|---|---|---|
| word2vec-word2vec | 90.43% | 76.52% |
| (True positives/Total) | (104/115) | (88/115) |
| word2vec-GloVe | 90.43% | 60.86% |
| (True positives/Total) | (104/115) | (70/115) |

**Table 5.** Results on Quora dataset

From the results, we observed that our model trained on word2vec alone, adapted better to the unseen real world examples than the one we trained with word2vec and GloVe. This could be due to the range of vocabulary in word2vec is high (1.2M) compared to GloVe (400K).

Finally we performed a similar experiment with two different GloVe models which are learned from different datasets and used them to create a multichannel CNN. The idea was that additional information from distinct vector spaces would add more context to each word, which in turn should yield higher accuracies. But we did not achieve accuracies higher than those achieved with single channel. Our multichannel model was overfitting and unable to utilize this extra contextual information to classify the questions more accurately.

## 8. CONCLUSION AND FUTURE WORK

In this work, we extend the existing method of convolutional neural networks for the task of question classification at two levels. Instead of using basic representations of text, we used the state-of-the-art vector representations such as word2vec and GloVe. We proposed an extended CNN architecture that can first classify a question into a broader category, and based on the prior knowledge, can classify it into a more specific category. We ran various experiments with pre-trained word2vec and GloVe models, and we tested it on two different datasets. On the TREC set, the primary accuracy was on par with the current methods, while the secondary accuracy improved compared to Pota et al. (2015) [2]. To test the robustness of the system, we tested the model on the dataset prepared by randomly selected questions from Quora. We tried to work with the idea of multi-channeling, but the results were not satisfactory. Therefore, in the future, we would like to extend our model upon that idea by creating a CNN that can properly capture the contextual importance of multi-channel representation. We would also like to train the model on new kinds of questions, that are not covered in our training data, to improve the robustness of our model.

## 9. ACKNOWLEDGMENTS

This project was a part of the Machine Learning for Signal Processing (E-599) course. We would like to thank Professor Minje Kim for his support and guidance during the course.